# Reproducible AI Requires Reproducible Randomness

## When Identical PRNG States Do Not Produce Identical Random Streams

**Anthony Bertrand, Tom Schmitt, Engelbert Mephu Nguifo, David R.C. Hill**

Université Clermont Auvergne, Clermont Auvergne INP, CNRS, LIMOS, F-63000 Clermont–Ferrand, France

anthony.bertrand@uca.fr [1]

**Abstract:**

*Pseudorandom number generators (PRNGs) constitute indispensable computational tools across multiple scientific domains, including Monte Carlo simulations, stochastic computing, and artificial intelligence (AI). The reproducibility of such applications critically depends on the ability of PRNG implementations to generate identical sequences across software environments when initialized from the same internal state. These algorithms enable the simulation of stochastic processes while providing deterministic and repeatable behaviour, thereby facilitating reproducible experiments. Modern PRNG implementations may be initialized through either a seed or, more accurately, an initial state that exceeds the capacity of a conventional integer seed. However, reliance on a simple seed alone frequently proves insufficient to ensure consistent program execution traces across different implementations. A natural assumption is that transferring the complete internal state of a generator should guarantee identical outputs regardless of the software library used. This study examines the validity of this assumption by investigating whether complete initial states can ensure cross-library fidelity and portability of PRNG streams. We focus on two widely deployed generators, Mersenne Twister and Philox, and evaluate their implementations across four major Python ecosystems—Random, NumPy, PyTorch, and TensorFlow. We compare the sequences produced by these implementations against those generated by the original reference algorithms under identical initialization conditions. Our results demonstrate that reproducibility cannot be assumed from PRNG state transfer alone, even when implementations claim to follow the same underlying algorithm. While fidelity was successfully achieved for several implementations, significant discrepancies were observed in others. Most notably, the Philox implementation in PyTorch exhibits fundamental incompatibilities with the reference algorithm, preventing exact reproduction of generator outputs across environments. These findings challenge the common expectation that access to a full internal state of a PRNG is sufficient to ensure reproducibility across software stacks. They further highlight that implementation-specific design choices can introduce hidden barriers to experimental replication, particularly in AI workflows that rely on multiple frameworks. This work shows that implementation fidelity of a PRNG is a necessary condition for scientific reproducibility and makes two primary contributions. First, it identifies practical guidelines for achieving reliable PRNG usage and reproducibility within the Python scientific and AI ecosystem. Second, it evaluates the extent to which cross-library portability and fidelity can be recovered through user-level techniques, without requiring modifications to library source code.*

---

[1] Contacting author.

## 1 Introduction

Pseudorandom number generators (PRNGs) constitute a fundamental component in various scientific disciplines. These algorithms facilitate the generation of random numbers in a deterministic and reproducible manner. Their applications are particularly prevalent in stochastic algorithms and experimental frameworks, such as Monte Carlo (MC) simulations. Furthermore, the field of artificial intelligence (AI) also benefits from these random number sequences, with applications ranging from data shuffling procedures to optimization of machine learning (ML) models, as an example, via stochastic gradient descent. Consequently, a comprehensive understanding of contemporary PRNG methods and their practical applications is essential for researchers and practitioners within the ML domain.

Library developers often implement PRNGs in alternative programming languages rather than utilizing their default implementations. This process, known as porting a library to a different environment, is exemplified by the Mersenne Twister (MT) algorithm, which was originally implemented in the C programming language, and adapted for use in Python through libraries such as Random, NumPy [1], and PyTorch [2]. It is natural to expect a PRNG to yield the same behaviour—the same random sequences—when identically initialized in different environments. However, a study investigating the performance of PRNGs [3] observed discrepancies in the generated sequences across different platforms when generators were initialized using identical seeds.

It is important to note that contemporary PRNGs do not exhibit a bijection between a seed and the complete generator state. The observed discrepancies in the previously mentioned study can easily be attributed to the initialization of generators using a seed rather than their full state. We suppose that the fidelity of these generators with respect to their original algorithmic specifications is compromised when initialized solely from a seed. Furthermore, we question the portability and fidelity of PRNGs across different libraries, particularly in the context of Python, which is extensively utilized in ML applications. Specifically, we wonder if it is feasible to guarantee fidelity between different libraries—i.e., generating identical sequences of random numbers—when utilizing the same PRNGs initialized through the library application programming interface (API), or even when employing the full generator state, given that seeding alone may be insufficient. These inquiries are important for the design and debugging of scientific experiments.

Our objective is to identify potential pitfalls arising from misunderstandings in code implementation and insufficient documentation. We will show that PRNGs are frequently misused due to three primary factors: (1) insufficient technical knowledge among researchers utilizing PRNGs, (2) ambiguous function nomenclature or documentation, and (3) erroneous implementations. The first point is addressed through the explanation of this article, while the other points are addressed with examples in Python libraries widely used in ML. Furthermore, we aim to highlight the significance of PRNGs within their respective scientific domains and to help users avoid the pitfalls we have experienced. Additionally, we emphasize the crucial importance of reproducibility in experimental scientific research.

In this investigation, we examine two prominent PRNGs: MT and Philox. The original MT algorithm, first introduced in 1998, continues to enjoy widespread adoption despite its well-documented statistical deficiencies, particularly its unsuitability for cryptographic applications. Mastsumoto, Saito and Nishimura propose CryptMT for this purpose [4]. In contrast, Philox is a counter-based PRNG that employs cryptographic functions to generate random sequences while offering inherent parallelization capabilities.

Our primary research objective is to achieve cross-library consistency in generated random sequences by initializing PRNGs with identical states. We hypothesize that certain library implementations may deviate from their original algorithmic specifications. The article is organized as follows: Section 2 defines the scientific fields where PRNG utilization is essential, with particular emphasis on AI (and more particular ML) applications. Section 3 highlights the critical importance of robust PRNG implementations. Section 4 addresses the importance of reproducible scientific research while providing relevant definitions. Section 5 enumerates the PRNGs selected for this study. The subsequent section documents our attempts to generate consistent random sequences across different libraries, highlighting potential implementation pitfalls and misconceptions regarding function behaviours. Section 7 presents a critical analysis of these results. Finally, the concluding section proposes a method for achieving sequence consistency when feasible, followed by our concluding remarks.

## 2 Random numbers in science with a focus on AI

The historical development of random number generation in scientific applications initially relied upon physical devices, such as dice, for producing stochastic outcomes. Early scientific literature reported the collection of random digits. We can cite Tippett [5] in 1927, considered to be the first, who gathered random numbers from many different census reports. Subsequent research expanded these efforts, such as Fischer [6] who published a comprehensive table of random numbers specifically tailored for applications in biological, agricultural, and medical research. Kendall [7] developed statistical tests, alongside a table of 100,000 random numbers. The RAND Corporation published in 1955 a book of one million digits through the simulation of an electronic roulette wheel [8].

Reading numbers from these sources was slow and memory inefficient. This limitation spurred the development of two distinct methods for random number generation [9]: (1) physical devices that exploit stochastic noise sources, and (2) deterministic algorithmic approaches. The first one gave rise to True Random Number Generators (TRNGs) including Quantum Random Number Generators, used in cryptography where the statistical independence of generated numbers is crucial, i.e. where the prediction of subsequent numbers from previously observed values must be impossible. The second method led to the development of Pseudo Random Number Generators (PRNGs).

PRNGs are specifically engineered to provide repeatable sequences, a critical feature for debugging and verification purposes in computational research. These generators are employed in scientific modelling and simulation contexts, where three fundamental requirements must be satisfied: (1) the capacity to trace program execution for proper experimental configuration; (2) the ability to reproduce identical digital experiments through consistent pseudorandom number sequences across multiple trials; and (3) the generation of statistically independent sequences across all experimental conditions to ensure unbiased statistical analysis.

An additional class of random number generators, though beyond the scope of this article, comprises quasirandom number generators. Unlike conventional PRNGs, these generators are not designed to produce sequences that appear statistically random. Instead, their primary objective is to achieve uniform distribution across the solution space [10]. Consequently, they fail conventional statistical randomness tests (elaborated in Section 3), yet their structured regularity confers distinct advantages for specific computational applications, including numerical integration, the design of experiments (DOE), and space-filling problems in low-dimensional contexts. These generators find particular utility in quasi-Monte Carlo simulation applications a set of specific applications where they often demonstrate superior convergence rates compared to traditional MC implementations utilizing PRNGs. They are useful for a limited number of dimensions. In addition, they eliminate the need for statistical replication and thus it saves computing time. Nevertheless, for most scientific application domains, PRNGs have remained the preferred methods for decades, primarily due to their operational simplicity and the critical advantage of repeatability they provide for debugging and for scientific research contexts.

MC simulations constitute a class of stochastic computational methods that employ random sampling techniques to approximate solutions for problems where analytical resolution proves either imprecise or mathematically intractable. Since their initial application, MC methods have undergone a paradigm shift from being considered a method of "last resort" to a state-of-the-art technique, primarily attributable to advancements in computational infrastructure and the relative simplicity of implementation when compared to certain mathematical approaches [11]. As MC simulations fundamentally constitute stochastic processes, they necessitate considerable quantities of random numbers—potentially on the order of hundreds of billions for a single physical experiment [12,13]. The resultant accuracy of these simulations exhibits direct dependence on the quality of the underlying

PRNGs, thereby establishing PRNG performance as a critical consideration in scientific applications [14].

Within the domains of AI and ML, PRNGs are present in the entire computational pipeline. During the data pre-processing phase, these generators facilitate numerous essential operations including: data augmentation [15] wherein transformations are applied to existing datasets to artificially expand their volume; data shuffling to mitigate biases arising from inherent ordering; data sampling to enable work with representative subsets of the complete dataset; and data splitting to partition datasets into training and testing subsets. During the model training phase, PRNGs serve critical functions in random initialization procedures, such as cluster center initialization in K-means clustering algorithms [16] and weight initialization in neural networks [17]. Additionally, they enable efficient data batching strategies that process subsets of training data to reduce computational overhead while enhancing model performance and generalization capabilities [18]. PRNGs also underpin optimization algorithms including stochastic gradient descent during backpropagation and the Adam optimization method [19]. Furthermore, they also appear in regularization techniques such as dropout wherein neurons are randomly deactivated to prevent model overfitting (defined as the phenomenon where models achieve excessive fidelity to training data at the expense of generalization to unseen data) [20]. Given their influence throughout AI and ML workflows, a comprehensive understanding of PRNG mechanisms and proper implementation strategies represents an essential competency for researchers seeking to maintain rigorous experimental control and reproducibility in their computational investigations.

## 3 Why statistically good PRNGs are important

### 3.1 Statistical properties of PRNGs

One of the earliest PRNGs was the middle-square method, introduced by John von Neumann in 1946. This approach proved inadequate due to its inherently limited cycle length. Subsequently, the linear congruential generator (LCG), proposed by Derrick H. Lehmer in 1949, significantly extended the length of random sequences through appropriate parameter selection. While LCGs can achieve favourable periodicity and one-dimensional uniformity, their structural deficiencies become apparent under more rigorous statistical tests. As demonstrated by George Marsaglia [21], all points generated by an LCG exhibit lattice structures, manifesting as parallel hyperplanes. Such regularities in random number generation frequently yield suboptimal outcomes in sophisticated MC simulation applications [22].

As Donald E. Knuth mentioned in his book [23]: “As computers become faster, more random numbers are consumed than ever before, and random number generators that once were satisfactory are no longer good enough for sophisticated application in physics, combinatorics, stochastic geometry

etc…". Consequently, statistical testing methods emerged to evaluate the quality of random number generators. While the mathematical definition of true randomness remains elusive, these tests effectively identify non-random characteristics by detecting patterns or dependencies within generated sequences. Numerous statistical properties can be examined through empirical validation procedures, including the frequency test to check that numbers generated are uniformly distributed, the serial test for examining consecutive number relationships, and the gap test for analysing interval distributions between specific value occurrences. It is important to acknowledge that absolute verification of true randomness remains unattainable; however, these tests provide probabilistic confidence in the randomness hypothesis of a given sequence.

In [24], L'Ecuyer enounced several essential qualities for PRNGs. These include an extended period length, computational efficiency, repeatability, and portability (portability is discussed in Section 4). Many of these characteristics can be quantitatively assessed through standardized test suites, such as the Diehard battery developed by George Marsaglia in 1996, its enhanced variant Dieharder by Robert G. Brown, and the comprehensive TestU01 framework by L'Ecuyer himself [25]. While these tests increase confidence in the randomness properties of a generator, they cannot guarantee absolute infallibility. In [26] Lawrence E Bassham gives statistical tests for PRNGs used in cryptographic application developed by the National Institute of Standards and Technology (NIST). Frederick James has made a review on high-quality PRNGs based on the theory of Mixing in classical dynamical systems suitable for MC simulations [27].

### 3.2 Impact of good PRNGs in ML

Regarding research on PRNG quality, Issah Zabsonre Alhassan conducted a comprehensive review in [28] comparing conventional PRNGs with chaos-based and ML-driven PRNGs. The study revealed that while chaos-based and ML-driven approaches can achieve superior statistical randomness, as measured against NIST cryptographic standards, they do so at the expense of significantly reduced generation speeds (typically $10^2$ to $10^3$ times slower than traditional methods). In [29], Nandapalan explored the implementation of PRNGs on graphics processing units (GPUs), leveraging their computational power and inherent parallelization capabilities.

Regarding the impact of PRNGs on result quality, Ferrenberg [30] demonstrated that high-quality random numbers "may lead to subtle, but dramatic, systematic errors for some algorithms". The article enounces the need to evaluate the interaction between specific algorithms and PRNGs, regardless of whether a generator passes statistical tests. Benjamin Antunes [31] quantitatively assessed the influence of statistically robust PRNGs on the outcomes of stochastic experiments. David Picard trained an identical AI model multiple times, varying only the initialization seed across experiments [32]. Although the specific PRNG implementation remained unspecified, the research context suggests the utilization of Philox (a state-of-the-art modern generator) given its deployment in PyTorch with GPU acceleration. The findings revealed that, under worst-case conditions, variations in seed selection could result in accuracy score discrepancies approaching 2%, a difference potentially significant for ML practitioners. Complementing these investigations, Hana Ahmed [33] analysed the performance variations of different ML algorithms based on the seed selection. The results indicated inconsistencies attributable to multiple factors, including algorithmic architecture, dataset characteristics, and train-test partitioning strategies. Koivu [34] examined the correlation between randomness quality and node dropout regularization in mitigating neural network overfitting. While the study identified potential relationships between randomness quality and model generalization capabilities, the observed effects were found to be highly contingent upon specific dataset properties.

## 4 Reproducibility

Reproducibility constitutes a pillar of experimental sciences. It is defined as the capacity for independent researchers to obtain identical results and scientific conclusions when reproducing a previously published study using all specified components of the original investigation [35]. This implies that the researcher must use the same data, experimental protocol, code, environment and all other relevant experimental parameters. Achieving reproducibility requires rigorous documentation of every aspect of the research process.

Faced with the reproducibility crisis that started around 2010 [36], initiatives have emerged to highlight its importance and improve the quality of scientific contributions [37]. The field of AI faces reproducibility challenges as demonstrated by Gundersen [38] which revealed that many papers published at renowned AI conferences were not fully reproducible. The primary causes include inadequate documentation and the absence of essential data and code artefacts. However, certain reproducibility challenges stem from more fundamental repeatability issues. Repeatability refers to the capability of a researcher to obtain consistent results when executing the same experimental procedure multiple times under identical conditions. Given that computational experiments are performed on deterministic hardware systems (classical computers), we expect achieving bitwise identical results, a level of repeatability essential for effective program debugging. The inability to properly debug computational experiments fundamentally undermines our capacity to conduct rigorous scientific investigations. The modern software and hardware stacks employed in AI research introduce additional complexities that may compromise repeatability through the introduction of stochastic elements into computational processes [39]. The inherent opacity and intricate interdependencies of contemporary computational toolchains necessitate careful consideration by researchers to ensure that their experimental results maintain sufficient consistency for subsequent verification by other scientific teams. Only through such comprehensive attention to reproducibility and repeatability can AI research achieve the standards required for scientific validity.

Repeatability is crucial for interpreting and explaining ML results because randomness in the training process can introduce variability that obscures the true performance of a model. For instance, consider a researcher training a neural network where randomness is introduced in three keyways: data shuffling, weight initialization, and dropout. After running the training process multiple times, the researcher observes that most runs yield test accuracies between 91% and 94%, but one run results in a significantly lower accuracy of 74%. This anomaly raises critical questions: Was it caused by an unfortunate data shuffle, a poor weight initialization, or the dropout mechanism? Since computers are deterministic, the issue must stem from the random elements in the training process. Without the ability to repeat the experiment under controlled conditions, the researcher cannot systematically isolate the cause to investigate and explain this discrepancy. This is a prerequisite for improving the reliability and interpretability of the model.

Researchers typically initialize PRNGs using a seeding mechanism, which ensures repeatable results but may prove insufficient for comprehensive reproducibility. Contemporary PRNG implementations possess internal states of such complexity that a simple integer seed cannot adequately represent them. For instance, the MT algorithm, as discussed later in Section 5, requires a minimum of 624 integer values plus an index to fully describe its internal state, occupying approximately 2 KB of binary data. With this example, it becomes evident that no bijective relationship exists between the potential seed values and the possible internal states of the generator. The specific implementation of the seeding function introduces further complications, as identical seed values may map to distinct internal states across different library implementations. Consequently, researchers are advised to manipulate and store the complete internal state of PRNGs rather than relying solely on seed values. Theoretically, this approach should ensure consistent generator behaviour even when implementations vary. However, this assumption constitutes the central research question of our investigation: Does the utilization of complete internal states genuinely guarantee identical generator behaviour across diverse implementations?

Portability constitutes a critical prerequisite for ensuring the replicability of MC simulations and ML studies across diverse computational environments, as defined by the formal characterization of ACM replicability[1]. While the concept of reproducibility enjoys well-established formal definitions, the term portability lacks an equivalent ACM-sanctioned definition. Poole and Waite [40] provided an early characterization in 1973, describing portability as "a measure of the ease with which a program can be transferred from one environment to another […]". As Gentle stated in [41], a PRNG producing high quality results in one computer-compiler environment will be as reliable in another environment. Portability allows experiments to be reproduced elsewhere and, consequently, allows researchers to expand their research. However, **the primary concern is fidelity**. How can a program be considered portable when it does not work as expected? The fidelity criterion in the definition of portability seems necessary.

[1] https://www.acm.org/publications/badging-terms

To start from the characterization above, a formal definition of portability could be “A measure of the ease with which a program can be transferred from one environment to another, while keeping the behaviour it has been created for, and the results it is supposed to produce”. A program P running on an environment E and giving an output O is **easily portable** on another environment E’ if we can obtain a program P’ with as little modification as possible (or no modification at all), and still produce the output O, within a stated precision. In this article, we discuss implementation fidelity and results fidelity of PRNGs. **We consider a PRNG to be portable if it produces the same output as that expected from the original implementation.** This definition will be used for the remaining of the article. When a PRNG generates divergent sequences of random numbers across different computational environments, we lose the capacity to reproduce the work of others. Under such conditions, this “fake portability” fails to guarantee meaningful scientific replications.

Portability issues raised here are (1) implementation fidelity and (2) modularity. First, the concept implementation “fidelity” concerns the degree to which a given PRNG implementation adheres to the original implementation. As previously discussed, mathematicians and statisticians incorporated rigorous statistical validation protocols to ensure the production of high-quality random sequences. Prominent PRNG researchers, including Makoto Matsumoto [42], have provided reference sequences with defined initial states to facilitate verification of reproducibility across different computational environments. When implementation-specific decisions result in divergent behaviours, the statistical integrity of the generated sequences may be compromised, thereby undermining the fundamental purpose of the PRNG.

Second, modularity refers to the requirement that multiple implementations of the same PRNG algorithm—particularly those employed across different programming languages—should exhibit consistent baseline behaviour while permitting performance optimizations and additional functional enhancements. Although various implementations may incorporate supplementary features such as post-processing transformations or normalization procedures, these modifications should not alter the core algorithmic behaviour. The implementations should be functionally interchangeable in terms of their fundamental operation, with performance variations arising solely from algorithmic optimizations rather than from modifications to the underlying random number generation process.

The present investigation builds upon prior research by Benjamin Antunes [43], who conducted a comprehensive assessment of PRNG statistical quality and reproducibility when deployed within ML frameworks relative to their original C implementations. His findings revealed significant discrepancies that raise concerns regarding implementation fidelity. Similarly, Hana Ahmed [44] examined portability challenges specifically concerning the MT algorithm across Python and C++ implementations—two important programming languages in ML research. Our work extends Ahmed's analysis by incorporating the Philox PRNG, a generator frequently utilized in AI applications. Furthermore, we investigate the mechanisms through which portability may be compromised and evaluate potential user-level strategies for mitigating such inconsistencies when feasible.

## 5 Studied PRNGs used in ML

To conduct an analysis of PRNGs within the context of ML applications, we first identified the specific generators employed by prominent ML libraries. Several widely adopted libraries rely on NumPy for their random number generation infrastructure, which incorporates the PCG algorithm as its default generator. However, deep learning frameworks such as TensorFlow [45] and PyTorch [2] implement their own dedicated PRNG solutions rather than utilizing the default generator of NumPy.

Documentation indicates that TensorFlow employs the Threefry and Philox algorithms, whereas PyTorch incorporates both the MT and Philox generators. Table 1 presents a comprehensive summary of the PRNG implementations across these ML libraries.

*Table 1 Summary of PRNGs used for each Python library.*

| Libraries \ PRNGs | MT | Philox | PCG | ThreeFry |
|---|---|---|---|---|
| Random | X | | | |
| NumPy | X | X | X | |
| PyTorch | X | X | | |
| Tensorflow | | X | | X |

As previously mentioned in a previous article [46], MT exhibits statistical flaws that render it unsuitable for cryptographic purposes. This problem does not affect simulation nor ML experiments. In contrast, the PCG algorithm demonstrates significant vulnerabilities in parallel computational contexts and exhibits predictable output characteristics[1], thereby rendering it inappropriate for both cryptographic applications and parallel stochastic experiments that require multiple independent random sequences. This assessment has been formally incorporated into the NumPy documentation. Furthermore, the proposed extensions to PCG have been demonstrated to exhibit failures, as documented in [46]. Threefry being only used by TensorFlow, we decided to study MT and Philox, as they are used in many well-known ML Python libraries.

[1] https://pcg.di.unimi.it/pcg.php

MT [42] represents a class of PRNGs that leverages Mersenne prime exponents to produce sequences characterized by an exceptionally large period of $2^{19937}-1$ and a 623-dimensional equidistribution property. While multiple variants within the MT family demonstrate enhanced statistical properties and computational efficiency, the original MT implementation remains the most prevalent in scientific applications. This algorithm serves as the default random number generator in the Python programming language and is incorporated into the C++ Standard Library. For parallel computational applications, two principal methods enable the utilization a PRNG [47]: partitioning a generator stream into smaller ones (substreams) and the creation of multiple generator with different parameters (multistreams). For MT, the dynamic creation technique [48] ensure the creation of multiple independent streams by using a unique identifier for each stream, that will be part of the statuses of the generators. The substream method can be handled by techniques like the random spacing, where we create random statuses for MT. With its very high period, we can easily find non-overlapping substreams of random numbers [22]. Many implementations are available: a 64-bit version, a SIMD-oriented Fast Mersenne Twister [49] or even a GPU-adapted version [50]. We will stick to the original Python implementation of MT, which is the original 32-bit version. It uses a state of 624 * 32-bit integers plus a 16-bit integer (around 2.4 KB).

Philox represents a member of the counter-based PRNG family [51], constructed upon cryptographic primitives for random bit generation. In contrast to MT, which maintains and iteratively updates a state vector to produce random numbers, Philox employs a key-counter mechanism. The period of Philox exhibits direct dependence on the counter bit-width. A 128-bit counter configuration enables the generation of up to $2^{128}$ pseudorandom 128-bit values (equivalent to 4 x 32-bit values), representing a long period suitable for most sequential computational applications. Philox is fast and memory-efficient, requiring storage of only the key and counter components (24 bytes for the 4 x 32-bit variant).

Furthermore, Philox exhibits particular suitability for GPU architectures, as it natively supports both multistream and substream parallelization paradigms. The multistream method instantiates distinct generator instances through key diversification in parallel execution contexts, whereas the substream paradigm exploits the extended period to partition the generator into multiple independent subsequences, each characterized by a reduced period. Substream generation presents minimal computational overhead, as the counter-based architecture enables arbitrary positioning within the pseudorandom sequence. For instance, a 128-bit counter configuration permits partitioning of the generator into $2^{64}$ independent substreams, each capable of producing $2^{64}$ pseudorandom values. Each substream maintains identical algorithmic behaviour while initiating its counter with a stride of $2^{64}$ sequence positions. This architectural characteristic may impose constraints on applicability within exascale supercomputing and large-scale AI cluster environments.

# 6 Experiments

In this section, we tried to achieve identical execution traces (defined as identical sequences of generated pseudorandom numbers) across different implementations of the same PRNG algorithms. For the purposes of this analysis, we designate the original Mersenne Twister implementation as the OMT and the original Philox implementation as OPX. Experimental results were stored as NumPy arrays, with a formatting applied to display only five significant digits for floating-point values to facilitate human readings. You can find more details in our pedagogical Jupyter Notebooks in our Gitlab repository[1].

Execution has been done on a Xeon Platinum 8470 CPU and a NVIDIA H100 GPU. We used Python 3.11.2 with TensorFlow 2.20.0 NumPy 2.3.2 and PyTorch 2.8.0+cu126.

OMT has been retrieved from Makoto Matsumoto's personal page[2] and OPX from the official repository[3].

## 6.1 OMT experiments

In this experimental phase, we used the OMT implementation as a reference for generating pseudo-random number sequences. The initial results obtained with the OMT, using a seed value of 42, are presented in Table 2. The OMT offers several generation functions; we focus our analysis on three specific functions commonly used: (1) *genrand_real2()* generates floating-point numbers uniformly distributed within the interval [0, 1), utilizing the underlying integer generation mechanism of the algorithm; (2) *genrand_res53()* constructs double-precision floating-point numbers by combining two consecutive 32-bit integer outputs from the generator, thereby achieving enhanced precision in the resulting values. (3) *genrand_int32()* produces 32-bit unsigned integers distributed across the range [0, $2^{32}$-1]. Note that the floating-point generation functions ultimately rely on the underlying integer generation process. Specifically, the output of the *genrand_res53()* function, which requires two consecutive integer outputs, looks like the output of the *genrand_real2()* function with a jump in the generation and a deviation starting at the ninth decimal place (ignored here to facilitate our readings).

*Table 2 Numbers generated by the OMT seeded with 42. Used as reference values.*

| Type (function used) | Output | | | | |
|---|---|---|---|---|---|
| Floats (genrand_real2) | 0.37454 | 0.79654 | 0.95071 | 0.18343 | 0.73199 |
| Floats (genrand_res53) | 0.37454 | 0.95071 | 0.73199 | 0.59866 | 0.15602 |

[1] https://gitlab.limos.fr/anbertrand1/random_portability
[2] https://www.math.sci.hiroshima-u.ac.jp/m-mat/MT/MT2002/emt19937ar.html
[3] https://github.com/DEShawResearch/random123

| Integers (genrand_int32) | 1608637542 3421126067 4083286876 787846414 3143890026 |
|---|---|

We conducted comparative analyses between the reference output generated by the OMT and the pseudorandom number sequences produced by four distinct library implementations:

- **Random**: The built-in Python module providing functions for generating both integer and floating-point random numbers.
- **NumPy/random**: The native random number generation interface of NumPy, which serves as the most straightforward approach for random number generation. By default, this implementation utilizes the MT algorithm as its underlying generator.
- **NumPy/random.Generator**: Referred to as **NumPy V2** in the official documentation, this represents the recommended modern API of NumPy for random number generation, replacing the legacy interface.
- **PyTorch/random**: The random number generation system of PyTorch, which populates tensor structures with pseudorandom values when computation is done on CPU. For our experimental validation, we instantiated a one-dimensional tensor, subsequently converting it to a NumPy array to enable direct comparison with the reference sequences.

The results of our initial comparative assessment, including the specific function calls and their corresponding parameters, are documented in Table 3.

*Table 3 Numbers generated after "seeding" the corresponding generator to 42. Functions used to generate the numbers are provided. Similarities with the expected numbers are in bold.*

| Libraries and modules | random | numpy.random | numpy.random.Generator | torch.random |
|---|---|---|---|---|
| Functions | random() | random() | random() | rand() |
| Real | 0.63943<br>0.02501<br>0.27503<br>0.22321<br>0.73647 | **0.37454**<br>**0.95071**<br>**0.73199**<br>**0.59866**<br>**0.15602** | 0.54199<br>0.61967<br>0.05737<br>0.8119<br>0.86009 | 0.88227<br>0.915<br>0.38286<br>0.95931<br>0.39045 |
| Functions | randint(0, 2**32 - 1) | randint(0, 2**32) | integers(0, 2**32) | randint(0, 2**32) |
| Integer | 2746317213<br>1181241943<br>958682846<br>3163119785<br>1812140441 | **1608637542**<br>**3421126067**<br>**4083286876**<br>**787846414**<br>**3143890026** | 2327846034<br>3904886566<br>2661450408<br>1733955692<br>246401338 | **3421126067**<br>**787846414**<br>3348747335<br>2563451924<br>1914837113 |

The documentation of Python *Random* module specifies that the *randint(a, b)* function generates pseudorandom integers inclusively between endpoints a and b, thereby including both boundary values. This specification contrasts with the default behaviour of other library implementations, which typically exclude the upper bound in their integer generation functions.

We observe that only the default random number generator of NumPy produced sequences that precisely matched the reference output generated by the OMT. Notably, the integer generation mechanism of PyTorch produced two consecutive values that aligned with the OMT sequence, while skipping certain intermediate values from the original sequence during the generation process, making it incorrect. Given these observations regarding sequence fidelity, we proceed to investigate the impact of full-state initialization on generator behaviour across implementations, as this approach represents a more robust method for ensuring cross-library consistency in pseudorandom number generation. We remind you that the utilized state was generated by the OMT with an initial seed value of 42. Results are presented in

Table 4.

*Table 4 Numbers generated after initializing the generator with a state. Functions used to generate the numbers are provided. Similarities with the expected numbers are in bold.*

| Libraries and modules | random | numpy.random | numpy.random.Generator | torch.random |
|---|---|---|---|---|
| Functions | random() | random() | random() | rand() |
| Real | **0.37454**<br>**0.95071**<br>**0.73199**<br>**0.59866**<br>**0.15602** | **0.37454**<br>**0.95071**<br>**0.73199**<br>**0.59866**<br>**0.15602** | **0.37454**<br>**0.95071**<br>**0.73199**<br>**0.59866**<br>**0.15602** | 0.88227<br>0.915<br>0.38286<br>0.95931<br>0.39045 |
| Functions | randint(0, 2**32 - 1) | randint(0, 2**32) | integers(0, 2**32) | randint(0, 2**32) |
| Integer | 4083286876<br>670094950<br>669991378<br>249467210<br>3720198231 | **1608637542**<br>**3421126067**<br>**4083286876**<br>**787846414**<br>**3143890026** | **1608637542**<br>**3421126067**<br>**4083286876**<br>**787846414**<br>**3143890026** | **3421126067**<br>**787846414**<br>3348747335<br>2563451924<br>1914837113 |

Following full-state initialization, the reliability of pseudorandom number generation improved across implementations, though discrepancies persisted in specific contexts. While the Python Random module now correctly generated floating-point sequences, **its integer generation remained inaccurate**. Both NumPy generation methods (legacy and V2 API) achieved complete consistency with the reference implementation.

However, the random number generation of PyTorch continued to produce the same incorrect sequences documented in Table 3, indicating fundamental implementation flaws that persist even with full-state initialization. This persistent inconsistency suggests that the pseudorandom number generation mechanism of PyTorch may require architectural review to achieve reliable cross-platform reproducibility.

### 6.2 OPX experiment

OPX provides a reference validation file (kat_vector) containing the expected pseudorandom sequences for three standardized initialization schemes:

- **Init 0**: The generator is initialized with all zero values, where the key and counter are set to zero for the 4x32-bit variant (key=(0,0), counter=(0,0,0,0)).
- **Init F**: The generator is initialized with maximum 32-bit unsigned integer values (MAX=0xFFFFFFFF in hexadecimal notation), resulting in key=(MAX, MAX) and counter=(MAX, MAX, MAX, MAX) for the 4x32-bit configuration.
- **Init PI**: The generator is initialized using the hexadecimal digits of PI, with key=(0xa4093822, 0x299f31d0) and counter=(0x243f6a88, 0x85a308d3, 0x13198a2e, 0x03707344) for the 4x32-bit variant.

Table 5 compares the output sequences generated by both the 4x32-bit and 4x64-bit Philox variants against these reference sequences, as the implementation of NumPy exclusively supports the 64-bit configuration. This comparative analysis serves to assess implementation fidelity across different bit-width configurations of the Philox algorithm.

*Table 5 Numbers generated by OPX depending on the initialization. Used as reference values.*

| Variant \ Initialization | Init 0 | Init F | Init PI |
|---|---|---|---|
| Philox 4x32-10 | 6627e8d5<br>e169c58d<br>bc57ac4c<br>9b00dbd8 | 408f276d<br>41c83b0e<br>a20bc7c6<br>6d5451fd | d16cfe09<br>94fdcceb<br>5001e420<br>24126ea1 |
| Philox 4x64-10 | 16554d9eca36314c<br>db20fe9d672d0fdc<br>d7e772cee186176b<br>7e68b68aec7ba23b | 87b092c3013fe90b<br>438c3c67be8d0224<br>9cc7d7c69cd777b6<br>a09caebf594f0ba0 | a528f45403e61d95<br>38c72dbd566e9788<br>a5a1610e72fd18b5<br>57bd43b5e52b7fe6 |

We compared the output of OPX with three couples of Python library/module:

- **NumPy/random.Generator:** The NumPy framework provides the Generator class as the exclusive mechanism for instantiating a 4x64-bit Philox generator.
- **PyTorch/random:** When executing computations on graphical processing units, PyTorch employs a 4x32-bit Philox implementation for random number generation.
- **Tensorflow/random:** TensorFlow offers the capability to select between two counter-based PRNG algorithms: Philox and Threefry. Similar to PyTorch, TensorFlow fills tensor structures with pseudorandom values through its random number generation interface.

It is important to note that while these Python libraries provide seeding functions for Philox initialization, the OPX specification does not incorporate such functionality. Consequently, our experimental protocol exclusively utilized the complete initial state for generator initialization rather than relying on seed-based approaches. The results of our initial comparative assessment, including the observed discrepancies between implementations, are documented in Table 6.

*Table 6 Numbers generated after initializing the generator with the corresponding state. Functions used to generate the numbers are provided. Similitudes with the expected numbers are in bold.*

| Libraries and modules | Init / Functions | Init 0 | Init F | Init PI |
|---|---|---|---|---|
| numpy.random.Generator | integers(0, 2**32) | 02f4ba6408e4d89b<br>3dd62b0b9ca8c5b2<br>1c8667a55d902e79<br>907d7a052fd5b4dc | 44b7493d1acfc229<br>6636af8e997921dd<br>3f73e132b5b3780e<br>605644dde03b01b1 | 4c8e672094922aa3<br>527061cd2884102a<br>f4c265b2d783d553<br>0556e76cb0298c8d |
| torch.random | randint(0, 2**32) | **e169c58d**<br>f08d6eaa<br>b6826759<br>115896b8 | None | None |
| tensorflow.random | uniform(0, 2**32) | **6627e8d5**<br>**bc57ac4c**<br>f8e4cca4<br>b1a574eb | **408f276d**<br>**a20bc7c6**<br>72a47709<br>9f41b01f | **d16cfe09**<br>**5001e420**<br>5757c6ce<br>3c0f08a0 |

NumPy generation is not correct. Torch generation can be done only for the Init 0 as it appears users do not have access to a part of the counter. For Init 0, the first generated number matches the second expected number. The rest is not correct. Tensorflow jumps some numbers from the expected sequence.

The experimental evaluation revealed significant implementation inconsistencies across Python library implementations of the Philox algorithm. NumPy produced sequences that did not match the OPX reference output, indicating fundamental flaws in its 4x64-bit Philox implementation. PyTorch demonstrates limited functionality. Users lack access to certain counter components. Consequently, we can only initialize the PRNG with the Init 0 initialization scheme. However, even this initialization presents a fundamental flaw, as only the first generated value corresponds to second expected value in the reference sequence. All subsequent numbers deviate from the expected output. Tensorflow similarly failed to reproduce the reference sequence, skipping multiple values in the expected output.

These findings collectively demonstrate portability and fidelity issues in current Philox implementations across major Python ML libraries, with each implementation exhibiting unique failure modes that compromise cross-platform reproducibility. These issues will be examined in order to identify their possible causes and find workarounds.

# 7 Discussion

Table 7 presents a comprehensive summary of our initial investigation into PRNG portability across different Python implementations. For the MT algorithm, the Python Random library employs a distinct seeding mechanism compared to the OMT. While this divergence could potentially introduce complications, the issue may be mitigated through the utilization of complete initial states rather than seed values to reproduce the reference generator output. This approach similarly applies to the implementation of NumPy V2.

PyTorch, however, exhibits atypical behaviour in integer generation. Notably, omitting one number between each generation yields sequences that align with the reference output. This observation suggests that the underlying mechanism correctly generates values but consumes two integers for each requested output, a behaviour that may introduce confusion for users. Surprisingly, the floating-point generation mechanism demonstrates different characteristics, as it does not exhibit this skipping behaviour.

Regarding the Philox algorithm, TensorFlow manifests the same "jump" phenomenon observed in the integer generation of PyTorch, where two numbers are consumed to produce a single output. In the case of the Philox implementation of PyTorch, we identified a fundamental architectural issue. The theoretical specification requires a 64-bit key and a 128-bit counter (typically represented as 2x32-bit key and 4x32-bit counter components). However, the internal state representation of PyTorch utilizes an 8-bit tensor of size 16 (128 bits total). When these bytes are grouped into 64-bit values, the first 64 bits represent the key, leaving only 64 bits for the counter, effectively halving the required counter space. Despite extensive experimentation with various initialization schemes, we were unable to obtain the correct output sequences for the tested configurations.

*Table 7 Natural portability of MT and Philox for each library.*

| | MT (seed) | MT (state) | Philox (state) |
|---|---|---|---|
| Random (floats) | No | Yes | |
| Random (int) | No | No | |
| NumPy (float) | Yes | Yes | |
| NumPy (int) | Yes | Yes | |
| NumPy V2 (float) | No | Yes | |
| NumPy V2 (int) | No | Yes | No |
| PyTorch (float) | No | No | |
| PyTorch (int) | No | No | No |
| Tensorflow (int) | | | No |

A critical challenge in cross-library PRNG portability stems from the heterogeneous representation of internal states across different software implementations. The OMT maintains its state as an array comprising 624 * 32-bit unsigned integers supplemented by a 16-bit index (approximately 2.4 KB of memory). In contrast, PyTorch represents this state as a tensor composed of 5056 8-bit unsigned integers (approximately 5 KB), which incorporates the core state information along with additional undocumented metadata and padding. This non-trivial structural divergence necessitates extensive manipulation to achieve interoperability between implementations.

Furthermore, while most libraries provide getter and setter functions for state management, these interfaces exhibit fundamental limitations: they operate exclusively within their native library contexts. Specifically, a state vector obtained through a getter function cannot be directly utilized as input to a setter function in a different library implementation. This constraint imposes significant burdens on users, who must possess detailed knowledge of the internal state representation of the generator in each library to establish functional bridges between disparate implementations.

Compounding these technical challenges, documentation inconsistencies and inaccuracies further complicate cross-library portability. Section 8.1 provides a concrete example of such documentation deficiencies, wherein the provided specifications deviate from the actual implementation behaviour. These issues collectively underscore the obstacles to achieving reliable cross-platform reproducibility in pseudorandom number generation for scientific computing applications.

## 8 Curated code

This investigation categorizes the sources of non-portability in PRNG implementations into three distinct classes:

(1) **User-induced non-portability** arises from incorrect utilization of PRNG interfaces induced by users or by the documentation. For such cases, we identify specific pitfalls and provide guidance to prevent their occurrence

(2) **User-resolvable non-portability** represents inconsistencies that can be addressed through appropriate user-level interventions without requiring modifications to library source code. We demonstrate this scenario through practical examples illustrating how users may achieve portability.

(3) **Fundamental non-portability** encompasses irreconcilable discrepancies that cannot be resolved through user-level modifications. For these cases, we analyse the underlying conditions that necessitate source code modifications to achieve compliance with original generator specifications.

The following subsections present code examples for categories (1) and (2), followed by an examination of the conditions that necessitate case (3) interventions.

### 8.1 MT curated code

Upon examination of the source code implementation, we identified a fundamental flaw in the function *randint(a, b)* of the Random library. The official documentation specifies that this function returns a uniformly distributed random integer within the inclusive range [a, b]. However, implementation analysis revealed an unintended increment operation applied to the upper bound parameter b during execution.

Consequently, users attempting to generate 32-bit unsigned integers through the invocation *randint*($0, 2^{32}-1$), presuming this would yield the maximum representable value, encounter a deviation from expected behaviour. To obtain the correct maximum value of $2^{32}-1$, users must instead specify b = $2^{32}-2$, effectively compensating for the internal adjustment of the library. While this discrepancy stems from an implementation error rather than user misunderstanding, the resulting non-portability manifests as incorrect PRNG utilization. Therefore, this issue is appropriately classified within the first category of problems (1), representing user-induced non-portability.

Investigative analysis of NumPy V2 revealed a critical discrepancy between its default seeding mechanism and that of the OMT implementation. Although the library provides an undocumented function—*legacy_seeding(seed)*—capable of reproducing the original seeding behaviour, this functionality lacks formal documentation within the official API reference. Consequently, users attempting to initialize generators according to standard specifications may inadvertently employ an incompatible seeding protocol, thereby introducing portability inconsistencies. This particular issue is classified within category (1).

In the context of the integer generation mechanism of PyTorch, our examination revealed a fundamental architectural deviation from the expected behaviour. Direct generation of 32-bit unsigned integers consistently produces sequences that omit intermediate values, as previously documented. To circumvent this limitation, a pragmatic resolution involves requesting 64-bit integers and subsequently decomposing these values into two distinct 32-bit unsigned integers through manual bit manipulation. The implementation strategy for this correction is presented in Code 1. This adjustment enables the recovery of the correct pseudorandom sequence without necessitating modifications to the underlying library source code, thereby qualifying this issue for classification within category (2) as a user-resolvable non-portability.

```python
import torch
import numpy as np
SEED=42
INT64_MIN = -(1<<63)
INT64_MAX =  (1<<63) - 1
MASK32    = 0xFFFFFFFF

# Initialize generator on CPU for MT
torch_generator = torch.Generator(device='cpu')
torch_generator.manual_seed(SEED)
randoms = []

# Draw only 2 numbers as we will transform each 64-bit integer drawn will be 2 32-bit unsigned
integers
numbers = torch.randint(INT64_MIN, INT64_MAX, size=(1,2), generator=torch_generator,
dtype=torch.int64)

# Transformation process
for value in numbers[0]:
    value_arranged = value - INT64_MIN      # Remove the signed bit and reverse everything.
    b = np.uint32(value_arranged & MASK32)  # uint32 to get rid of negative numbers
    a = np.uint32(value_arranged >> 32)     # uint32 to get rid of negative numbers
    randoms.append(int(a))                  # back to int to remove data type in display
    randoms.append(int(b))                  # back to int to remove data type in display

# Display
print(f"randoms={randoms}")
print(f"Expected={MT_INTEGERS_REF[:4]}")

# Output:
# randoms =[1608637542, 3421126067, 4083286876, 787846414]
# Expected=[1608637542, 3421126067, 4083286876, 787846414]
```

*Code 1 PyTorch library: MT curated version for integer generation.*

Upon initial investigation, we hypothesized that the observed skipping behaviour during integer generation might represent a consistent pattern across all data types. However, an analysis revealed that the floating-point generation of PyTorch employs entirely distinct computational pathways compared to the OMT. While OMT follows a conventional approach of first generating 32-bit integers and subsequently applying normalization through division by $2^{32}$ to obtain values in the [0, 1) interval, PyTorch implements a fundamentally different normalization paradigm.

Specifically, the implementation of PyTorch applies a 23-bit mask to the generated integer values prior to performing division by $2^{24}$ for normalization. The precise location of this normalization function within the source code of PyTorch remains undocumented and non-trivial to locate. Nevertheless, we were able to reconstruct this behaviour through empirical analysis of the operational characteristics of PyTorch, corroborated by community discussions documented in the PyTorch forum[1]. The reconstructed implementation strategy is presented in Code 2. This divergence between OMT and PyTorch normalization procedures represents a critical source of non-portability that cannot be addressed through simple user-level corrections, thereby necessitating category (3) classification for this particular implementation inconsistency.

[1] https://discuss.pytorch.org/t/how-does-torch-rand-sample-from-uniform-distribution/192544

```python
FLOAT_MASK = (1 << 24) - 1
FLOAT_DIVISOR = 1 / (1 << 24)

list_floats = []
for int_x in randoms_ints: # randoms_int = numbers of OMT
    list_floats.append((int_x & FLOAT_MASK) * FLOAT_DIVISOR)

print(np.array(list_floats[:5]))    # Reconstruction of the output of PyTorch from OMT
print(np.array(randoms[:5]))        # Output of PyTorch

# [0.88227 0.915   0.38286 0.95931 0.39045]
# [0.88227 0.915   0.38286 0.95931 0.39045]
```

*Code 2 Torch float generation according to a discussion on PyTorch website.*

Following this reconstruction of the normalization procedure, we confirmed that the floating-point generation mechanism of PyTorch precisely implements the 24-bit normalization paradigm previously described. This implementation choice results in the production of 24-bit floats rather than the expected 32-bit representations, fundamentally altering the statistical properties of the generated sequences. The only viable method to restore correct 32-bit floating-point generation involves manually constructing these values from the curated 32-bit integer sequences produced by our modified integer generation protocol. This particular case is classified within category (3) of non-portability, as the fundamental implementation contains intrinsic flaws that necessitate complete rewriting of the affected function to achieve compliance with the original algorithmic specification. Comprehensive implementation examples demonstrating this correction strategy are available in our supplementary Jupyter Notebook.

These interventions successfully resolve the identified portability inconsistencies, qualifying the observed issues for classification within category (2) as user-resolvable implementation discrepancies. The efficacy of these corrections is quantitatively demonstrated in Table 8 and Table 9, which present the improved portability metrics for all studied libraries following implementation of the proposed solutions.

*Table 8 Numbers generated with our curated version of the code after seeding the corresponding generator to 42. Functions used to generate the numbers are provided. Similitudes with the expected numbers are in bold.*

| Libraries and modules | random | numpy.random | numpy.random.Generator | torch.random |
|---|---|---|---|---|
| Functions | random() | random() | random() | See Notebook |
| Real | 0.63943<br>0.02501<br>0.27503<br>0.22321<br>0.73647 | **0.37454**<br>**0.95071**<br>**0.73199**<br>**0.59866**<br>**0.15602** | **0.37454**<br>**0.95071**<br>**0.73199**<br>**0.59866**<br>**0.15602** | **0.37454**<br>**0.95071**<br>**0.73199**<br>**0.59866**<br>**0.15602** |
| Functions | randint(0, 2**32 - **2**) | randint(0, 2**32) | integers(0, 2**32) | See Code 1 |
| Integer | 2746317213<br>478163327<br>107420369<br>3184935163 | **1608637542**<br>**3421126067**<br>**4083286876**<br>**787846414** | **1608637542**<br>**3421126067**<br>**4083286876**<br>**787846414** | **1608637542**<br>**3421126067**<br>**4083286876**<br>**787846414** |

| | 1181241943 | **3143890026** | **3143890026** | **3143890026** |
|---|---|---|---|---|

*Table 9 Numbers generated with our curated version of the code after initializing the generator with a state. Functions used to generate the numbers are provided. Similitudes with the expected numbers are bolded.*

| Libraries and modules | random | numpy.random | numpy.random.Generator | torch.random |
|---|---|---|---|---|
| Functions | random() | random() | random() | See Notebook |
| Real | **0.37454**<br>**0.95071**<br>**0.73199**<br>**0.59866**<br>**0.15602** | **0.37454**<br>**0.95071**<br>**0.73199**<br>**0.59866**<br>**0.15602** | **0.37454**<br>**0.95071**<br>**0.73199**<br>**0.59866**<br>**0.15602** | **0.37454**<br>**0.95071**<br>**0.73199**<br>**0.59866**<br>**0.15602** |
| Functions | randint(0, 2**32 - **2**) | randint(0, 2**32) | integers(0, 2**32) | see Code 1 |
| Integer | **1608637542**<br>**3421126067**<br>**4083286876**<br>**787846414**<br>**3143890026** | **1608637542**<br>**3421126067**<br>**4083286876**<br>**787846414**<br>**3143890026** | **1608637542**<br>**3421126067**<br>**4083286876**<br>**787846414**<br>**3143890026** | **1608637542**<br>**3421126067**<br>**4083286876**<br>**787846414**<br>**3143890026** |

### 8.2 Philox curated code

Following detailed examination of implementation behaviours, we identified that the generator of NumPy advances its internal counter prior to producing output values. By incorporating this counter incrementation into our reconstruction method, we successfully achieved sequence consistency with the reference implementation.

For TensorFlow, the reconstruction approach previously described in Section 8.1 proved effective. Specifically, we generate 64-bit integers and decompose them into two 32-bit unsigned integers to recover the expected sequence.

PyTorch presents fundamental portability limitations due to its parallel execution architecture. As discussed in Section 6.2, the PyTorch Philox state representation exposes only the key and the first 64 bits of the counter to user manipulation, while reserving the remaining 64 bits for internal parallelization management. The default Philox configuration partitions the generator into $2^{64}$ independent substreams, each capable of producing $2^{64}$ pseudorandom numbers—a configuration adequate for most applications but potentially insufficient for exascale computing environments. Here you will find a useful blog[1] post regarding the behavior of the Philox implementation in PyTorch.

When invoking the *randint()* function, PyTorch requires specification of the target tensor shape. To populate this tensor, the framework instantiates a number of threads (or jobs) equal to the number of required random values. Each thread maintains its own generator instance with a distinct counter value. The counter progression mechanism is graphically represented in Figure 1. For each state of Philox, we generate 128 bits (often split in 4 x 32 bits). In light of the explanation of the process below, questions arise: where are the remaining 96 bits of the generation? In addition, why do we only have access to the second 32-bit number?

[1] https://blog.codingconfessions.com/p/how-pytorch-generates-random-numbers

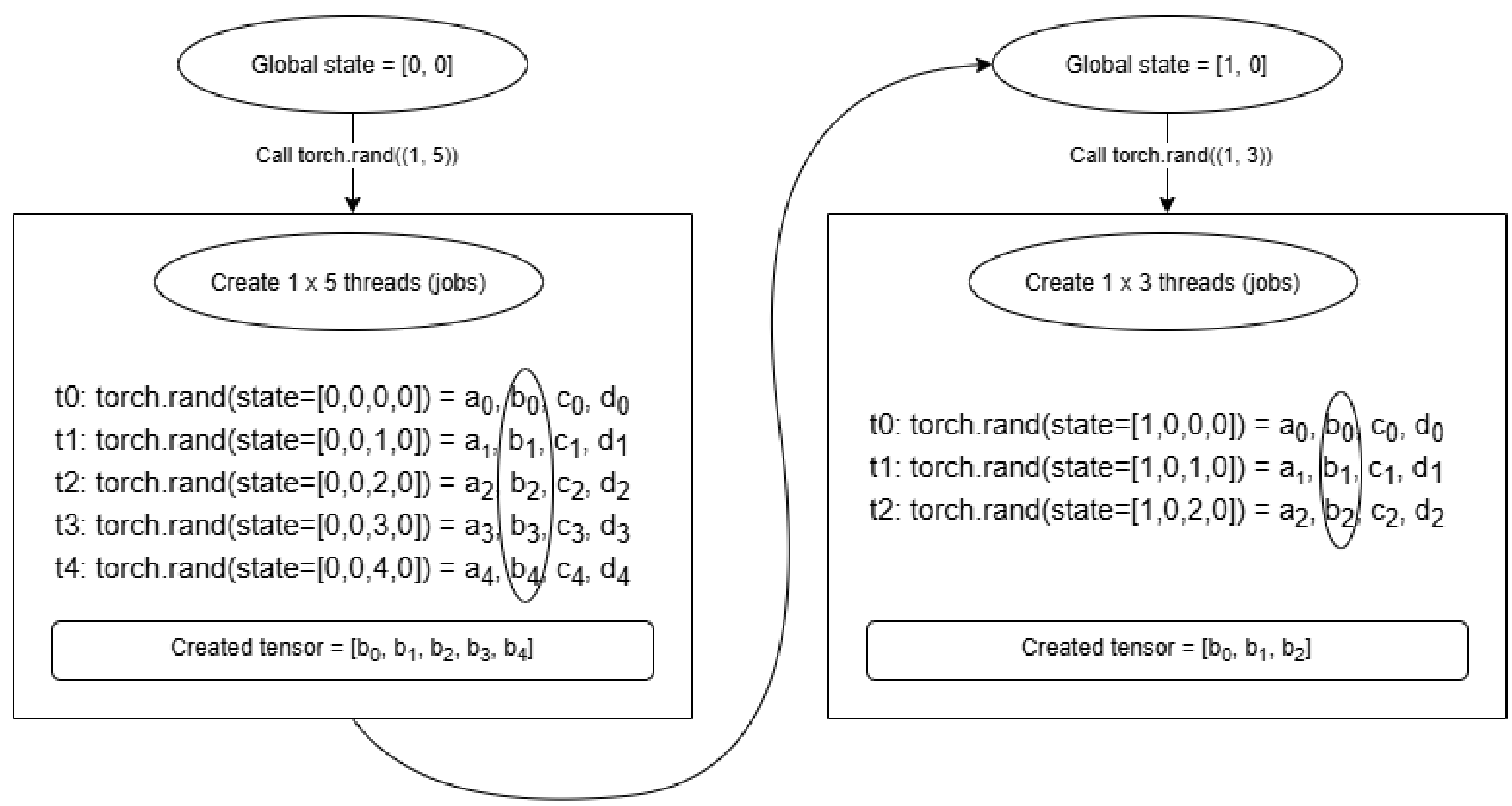

*Figure 1 Evolution of the PyTorch Philox counter. At each call of the rand() or randint() method, the global state is used to create n threads (n equal to the number of random numbers necessary to fill the tensor). Each thread has a state equal to the global state as the offset, and the thread index as the subsequence. For the first call in this example, the offset is [0,0] and the subsequence is [0,0], [1,0]…[4,0] to generate five numbers. For the second call, the offset is incremented by one. It is now [1,0] and the subsequence is [0,0], [1,0] and [2,0] to generate three numbers. Each thread generate one random number equal to the second 32-bit number generated by Philox.*

In addition to the counter management constraints previously discussed, the Philox implementation of PyTorch exhibits a fundamental architectural limitation in its number generation mechanism. While the algorithm is designed to produce four 32-bit integers per computational cycle, the library exposes only one of these four generated values to the user through its standard interface. This represents a significant departure from the expected behaviour of the OPX specification.

The most viable user-level workaround to recover the correct sequence involves generating tensors of shape (1,1) at each invocation. This approach effectively retrieves the intended pseudorandom values, though it introduces a skip of three subsequent numbers in the sequence between each generation.

The efficacy of these interventions is quantitatively assessed in Table 10, which presents the updated portability metrics for all examined library implementations following the implementation of the proposed corrections.

*Table 10 Numbers generated with our curated version after initializing the generator with the corresponding state. Functions used to generate the numbers are provided. Similitudes with the expected numbers are bolded.*

| Libraries and modules | Init / Functions | Init 0 | Init F | Init PI |
|---|---|---|---|---|
| numpy.random.Generator | integers(0, 2**32) | **16554d9eca36314c**<br>**db20fe9d672d0fdc**<br>**d7e772cee186176b**<br>**7e68b68aec7ba23b** | **87b092c3013fe90b**<br>**438c3c67be8d0224**<br>**9cc7d7c69cd777b6**<br>**a09caebf594f0ba0** | **a528f45403e61d95**<br>**38c72dbd566e9788**<br>**a5a1610e72fd18b5**<br>**57bd43b5e52b7fe6** |
| torch.random | randint(0, 2**32) | **e169c58d**<br>f08d6eaa<br>b6826759<br>115896b8 | None | None |
| tensorflow.random | uniform(0, 2**32) | **6627e8d5**<br>**e169c58d**<br>**bc57ac4c**<br>**9b00dbd8** | **408f276d**<br>**41c83b0e**<br>**a20bc7c6**<br>**6d5451fd** | **d16cfe09**<br>**94fdcceb**<br>**5001e420**<br>**24126ea1** |

# 9 Portability summary

Table 11 presents a comprehensive summary of the portability outcomes achieved following our correction of implementation inconsistencies across multiple MT and Philox implementations. While these interventions enabled the recovery of consistent pseudorandom sequences in certain contexts, a fundamental question persists regarding the practical feasibility of such corrections within authentic scientific research workflows.

Consider, for instance, the case of MT portability within PyTorch environments. To achieve sequence consistency with reference implementations, users must perform non-trivial tensor manipulation—specifically decomposing 32-bit tensors into their constituent components to reconstruct the intended MT generation sequence. Such manual intervention represents a deviation from standard scientific programming practices, raising significant concerns about the sustainability of these correction methods in rigorous research contexts. The practical implications of this requirement for manual sequence reconstruction within operational scientific workflows merit careful consideration and evaluation.

*Table 11 Progress made in PRNGs portability after using our corrections. (1) Means the generator works as intended, but can be wrongly used by the user. (2) Means a user can obtain the same random sequences through bit manipulations.*

| | MT (seed) | MT (state) | Philox (state) |
|---|---|---|---|
| Random (floats) | No | Yes | |
| Random (int) | No | Yes (1) | |
| NumPy (float) | Yes | Yes | |
| NumPy (int) | Yes | Yes | |
| NumPy V2 (float) | Yes (1) | Yes | |
| NumPy V2 (int) | Yes (1) | Yes | Yes (1) |
| PyTorch (float) | No | No | |
| PyTorch (int) | Yes (2) | Yes (2) | No |
| Tensorflow (int) | | | Yes (2) |

## 10 Conclusion

In this paper, we examined the portability and fidelity of Pseudo Random Number Generators, which constitute essential prerequisites for achieving reproducible scientific research. We formalized a more accurate definition of what portability should be, by adding the fidelity criterion, before analysing the portability of PRNGs. A part of the problem is that we hypothetically assess that a complete internal state initialization ensures identical output sequences across different software libraries. Our analysis focused on two widely adopted PRNG algorithms—Mersenne Twister and Philox—when deployed across multiple Python ML libraries. Theoretically, the internal state of a PRNG contains sufficient information to serve as a reliable initialization mechanism. However, our experimental results show that pragmatic implementation choices in library development invalidate this assumption. This discrepancy manifests in the floating-point random number generation of PyTorch, which implements a normalization procedure that diverges fundamentally from the canonical MT specification that the library purportedly utilizes. While PRNGs constitute indispensable computational tools for modelling stochastic processes, their improper implementation introduces errors that fundamentally compromise the integrity of research outcomes. We think we are using a 'good' generator *g* while we are using another implementation *g'* sometimes even with a different number of bits (e.g. PyTorch float generation). This affects particularly the domains of stochastic simulation and ML (and AI in general).

Through comprehensive empirical analysis, we established that PRNG portability across implementations could be categorized into three distinct profiles: (1) implementations that inherently produce identical execution traces; (2) implementations that require user-level interventions to achieve trace consistency; and (3) implementations where trace consistency is fundamentally unattainable without modification of the underlying source code. This taxonomy provides a practical framework for assessing the reproducibility guarantees offered by modern software libraries. Detailed computational notebooks accompany this investigation to enable reproducibility, facilitate independent verification, and provide actionable guidance for practitioners.

Portability represents a fundamental component of reproducible scientific research. The erosion of portability effectively constrains computational artefacts within specific execution environments, thereby compromising the capacity for independent verification of research findings. While contemporary development practices may prioritize computational efficiency or implementation convenience, such compromises fundamentally undermine the very foundation of empirical research. Our findings therefore raise an important question for the scientific software community: how much implementation freedom can be introduced before reproducibility itself is compromised?

The existing body of research has extensively evaluated the statistical characteristics of PRNGs through comparative analyses between different algorithms [14,25,26,43,52]. However, far fewer studies have investigated whether different implementations of the same algorithm preserve identical statistical behaviour and reproducibility guarantees across software ecosystems. While studies such as [32] and [33] investigated the impact of multiple seed values on ML training outcomes, the influence of complete initialization states and implementation-dependent behaviour remains largely unexplored. This research gap has been partially addressed by Wartel's comprehensive study [52], which demonstrated that modern PRNG implementations, when tested with thousands of distinct initialization states using Pierre L'Ecuyer's Big Crush test suite [25], exhibit statistical failures in approximately 30% of cases. Taken together, these findings suggest that evaluating PRNG algorithms alone is no longer sufficient. Reproducible scientific computing also requires evaluating initialization strategies and implementation fidelity. The conventional focus on identifying "good PRNGs" must be complemented by rigorous assessment of "good initializations" and appropriate parallelization methods, as emphasized by Hill's 2013 work [22]. Historical precedents reinforce this perspective: during the era of linear congruential generators (LCGs), the selection of statistically optimal parameters was not merely recommended but mandatory to mitigate inherent deficiencies and even these measures proved insufficient in isolation. Verifying the correctness of a PRNG algorithm alone is insufficient if software implementations cannot faithfully reproduce its behaviour across different computing environments. To facilitate future investigations and repeatability checks, we provide a curated repository[1] of statistically validated initialization states and corresponding reference sequences.

Given that the correlation between statistically validated PRNGs and optimal performance in stochastic experiments remains empirically unverified, we advocate complementing conventional PRNG evaluation with application-oriented validation protocols. Rather than assuming that a statistically validated generator will automatically yield reproducible scientific results, researchers should verify that both the selected implementation and its initialization strategy preserve the properties required by their specific application. Such validation protocols would strengthen both statistical confidence and experimental reproducibility across modern stochastic computing and AI workflows.

[1] https://gitlab.isima.fr/thwartel/testu01_variousgenerators

## Acknowledgements

I want to thank Fanta Sanogo for her work on Philox portability, and Tom Schmitt for his work on Mersenne Twister portability (https://gitlab.isima.fr/toschmitt/mt-pytorch-numpy-compatibility).

We used a Mistral AI LLM to improve grammar and overall comprehension without altering the intended meaning. The text prior to LLM revision has been saved and can be provided.

## Funding support

This work is financed by the CPER IDEAL and Clermont Auvergne Métropole (CAM).

## BIOGRAPHIES

**ANTHONY BERTRAND** is a PhD Student at Clermont Auvergne University (UCA) in the LIMOS laboratory (UMR CNRS 6158). He holds a Master in Computer Science (second head of the list). His thesis subject is about the software-based measurement of energy consumption of Machine Learning programs in High Performance Computing (HPC). He also addresses the reproducibility challenges in the domain of Machine Learning. His email address is anthony.bertrand@uca.fr.

**DAVID R. C. HILL** is a full professor of Computer Science at University Clermont Auvergne (UCA) doing his research at the French Centre for National Research (CNRS) in the LIMOS laboratory (UMR 6158). He earned his Ph.D. in 1993 and Research Director Habilitation in 2000 both from Blaise Pascal University and later became Vice President of this University (2008-2012). He is also past director of a French Regional Computing Center (CRRI) (2008-2010) and was appointed two times deputy director of the ISIMA Engineering Institute of Computer Science – part of Clermont Auvergne INP, #1 Technology Hub in Central France (2005-2007 ; 2018-2021). He is now Director of an international graduate track at Clermont Auvergne INP. Prof Hill has authored or co-authored more than 300 papers and has also published several scientific books. He also supervised research at CERN in High Performance Computing (https://isima.fr/~hill/).

**ENGELBERT MEPHU NGUIFO** is a full professor of computer science at University Clermont Auvergne (UCA), France, where he is the director of Master Degree Program in Computer Science. He is leading research on machine learning and data mining for complex data in the joined University-CNRS laboratory LIMOS where he is co-chair of the Information and Communication Systems research group. His research interests also include formal concept analysis, artificial intelligence, pattern recognition, bioinformatics, big data, and knowledge representation. He was Board member of the French Association on Artificial Intelligence. He is member of the executive board of the French CNRS research group on Artificial Intelligence (GDR RADIA).

**TOM SCHMITT** is a second-year bachelor student at Clermont Auvergne University (UCA). He is particularly interested in low-level programming and optimizations.

**ROLES**

**Bertrand Anthony**: Writing – Original Draft, Software, Testing

**Schmitt Tom**: Review & Editing, Software, Testing

**Mephu Nguifo Engelbert M.**: Review & Editing

**Hill David R.C.**: Writing – Review & Editing, Supervision, Validation